\documentclass[conference]{IEEEtran}
\IEEEoverridecommandlockouts
\usepackage{cite}
\usepackage{amsmath,amssymb,amsfonts}
\usepackage{algorithmic}
\usepackage{algorithm}
\usepackage{graphicx}
\usepackage{textcomp}
\usepackage{xcolor}
\def\BibTeX{{\rm B\kern-.05em{\sc i\kern-.025em b}\kern-.08em
		T\kern-.1667em\lower.7ex\hbox{E}\kern-.125emX}}

\title{Simulation-based Methods for Optimal Sampling Design in Systems Biology\\

}

	\author{\IEEEauthorblockN{Tuan Minh Ha}
	\IEEEauthorblockA{\textit{Department of Basic Sciences} \\
		\textit{Saigon Technology University}\\
		Ho Chi Minh City, Vietnam\\
  tuan.haminh@stu.edu.vn}
	\and
	\IEEEauthorblockN{Binh Thanh Nguyen}
	\IEEEauthorblockA{\textit{Department of Computer Science} \\
		\textit{University of Science, VNU}\\
		Ho Chi Minh City, Vietnam \\
		ngtbinh@hcmus.edu.vn}
	\and
	\IEEEauthorblockN{Lam Si Tung Ho}
	\IEEEauthorblockA{\textit{Department of Mathematics and Statistics} \\
		\textit{Dalhousie University}\\
		Halifax, Nova Scotia, Canada \\
		lam.ho@dal.ca}
	
}

\begin{document}

	\maketitle

\begin{abstract}
In many research areas of systems biology including virology, pharmacokinetics, and population biology, researchers often use the dynamical systems to describe dynamics of biological systems. 
We can learn about these biological systems by estimating the parameters of the dynamical systems from sampling data.
Therefore, an important question arises: how can one select the optimal sampling points for accurately estimating the parameters? 
Classical methods often rely on Fisher information matrix-based criteria such as A-, D-, and E-optimality.
However, these methods require an initial estimate of the parameters and often lead to a suboptimal result when the initial estimate is inaccurate.
In this paper, we develop two simulation-based methods for optimal sampling design that do not require an initial estimate of the parameters.
The first method, E-optimal-ranking (EOR), employs the E-optimal criterion while the second method utilizes the Long short-term memory (LSTM) neural network.
We demonstrate the performance of our proposed methods using simulation studies based on two popular models in system biology (Lotka–Volterra and Three-compartment models).
The results show that our methods outperform the naive random selection method and the classical E-optimal design.
\end{abstract}

\begin{IEEEkeywords}
optimal sampling design, systems biology, E-optimal design, Lotka-Volterra model, Three-compartment model, LSTM.
\end{IEEEkeywords}

\section{Introduction}
Dynamic system modeling is an essential tool for studying the dynamics of biological systems including the changes in the population of species (e.g. Lotka-Volterra models \cite{volterra1927variazioni,lotka1925elements,kot2001elements}), the transformation and metabolism of substances within the body (e.g. pharmacokinetic models \cite{khorasheh1999application,tam2003optimal}), and the fluctuation of virus levels within a living organism (e.g. HIV models \cite{stafford2000modeling,jo2019robust,adams2005hiv}). 
We often represent these dynamic models using systems of differential equations and learn the dynamics of biological systems by estimating the parameters from sampling data.
Therefore, designing an optimal sampling scheme is one of the most important tasks in systems biology.

Conducting experiments and sampling to estimate the parameters of a dynamic system is often time-consuming, labor-intensive, and costly. Therefore, Optimal Experiment Design (OED) is a crucial step in identifying a dynamic model. Classical criteria for optimal experimental designs are derived by optimizing the information matrix. These criteria include minimizing the trace of the inverse of the information matrix (A-optimal), maximizing the determinant of the information matrix (D-optimal), and maximizing the smallest eigenvalue of the information matrix (E-optimal). Several OED methods using these classical criteria have been proposed over the past decades \cite{d1981optimal,gil2014application,hagen2013convergence,yu2015optimal, paquet2015optimal}. 
Most of the works have a common drawback in that they require an initial estimate of the parameters and tend to obtain suboptimal results due to the inaccuracy of the initial estimate \cite{yu2018robust}.
Hence, it is essential to develop an effective experimental design method that does not require an initial estimate of the parameters. 

 To overcome this limitation of the classical optimal design criteria, we develop two simulation-based approaches for OED. 
 The first method adapts E-optimal criterion to avoid the need for an initial estimate of the parameters. The key idea is similar to Bayesian inference where we assume that the initial estimate follows a uniform distribution in the parameter space. Specifically, we simulate a vector of parameters uniformly from the parameter space and use E-optimal criterion to rank the sampling times according to the sampled parameters. We repeat this process several times and use the average ranking to pick the optimal sampling points.
  The second approach involves using a combination of simulated data and the Attention-based Long Short-Term Memory (At-LSTM) neural network to identify the sampling times that influence the estimation of parameters the most. 
 We illustrate our methods using two well-known dynamic models (the Lotka–Volterra and the three-compartment models).
 The results show that the proposed methods outperform the classical E-optimal criterion in selecting optimal sampling times for estimating the parameters of the dynamic models.

\section{Dynamic system modeling}

\subsection{Dynamic system models}

Dynamical system is a popular modeling tool in Systems Biology.
It has been used extensively to describe changes in the population size of organisms \cite{lotka1925elements,volterra1927variazioni}, the concentration of a substance \cite{tam2003optimal,khorasheh1999application}, cell signaling \cite{faller2003simulation}, synthetic biology \cite{liepe2013maximizing}, and infectious disease epidemics \cite{ho2018birth, saha2023spade4}. 
A general dynamic system model takes the form of a system of differential equations, which is typically structured as follows:
\begin{equation}\label{eqs:hedonghoctongquat}
	\begin{aligned}
		\dot{X}(t) &= f(X(t),\theta) \\
		X(t_{0})&=X_{0},
	\end{aligned}
\end{equation}
where $f(\cdot, \theta)$ is the functions of states transition which refer to the reaction mechanisms, $\theta \in \mathbb{R}^{m}$ is a parameter of the dynamic system, $X(t) \in \mathbb{R}^{n}$ is the state of the system at time $t$, and $X_0$ is the state of the system at time $t_0$ where we start observing the system. 

In practice, we may not be able to observe $X(t)$ directly due to measurement errors. Let $y\in \mathbb{R}^{q}$ be the measurement output vector, representing observation data. We have
\begin{equation}
    y(t) = g(X(t)) + w.
\end{equation}
Here, $g(\cdot)$ is the measurement function used to select measurement variables.  For instance, if all states of the system are measured, $g(\cdot)$ is an identity function. If we only observe the $i$-th state, then the function $g(\cdot)$ will be a projection function from the space $\mathbb{R}^{n}$ onto $\mathbb{R}$, keeping only the $i$-th component. The noise $w$ is a Gaussian random variable with zero mean.
 

In many studies, the main task is to estimate the unknown parameter $\theta$ of the system using the observations.
The least-square method is a commonly used method for this task. This estimator is expressed as:

\begin{equation}
	\begin{aligned}
		\hat{\theta} = \arg \min_{\theta \in \Theta}  \sum_{i=1}^{N} \|y(t_i) - g(X_\theta(t_i))\|^2
	\end{aligned}
 \label{LS}
        \end{equation}
where $\|\cdot\|$ is the Euclidean norm, $\Theta$ is a space of parameters, $X_\theta$ is the solution of the system of differential equations \eqref{eqs:hedonghoctongquat} with respect to $\theta$,  $N$ is the total number of observations, and $(t_i)_{i=1}^N$ are sampling times.
Throughout this paper, we assume that $\Theta$ is bounded.

\subsection{Classical Optimal Sampling Design}

The goal of classical optimal sampling design is to choose sampling points $t_i \quad (i = 1, 2, 3, \ldots, N)$ that maximize the information about the unknown parameter $\theta$. The Fisher information matrix (FIM) is used to measure the amount of data information contained in the experimental data. It can be built based on the parametric sensitivity matrix and the measurement error covariance matrix as follows 
\begin{equation}
	\text{FIM} = S^{T}S,
\end{equation}
where, $S = \partial X/\partial \theta$ represents the local parametric sensitivity matrix, which indicates how parameters locally influence model outputs \cite{brown2008robust}. $S$ can be solved using \eqref{eqs:hedonghoctongquat} and the following system of equations:
\begin{equation}\label{Matran S}
    \dot{S} = JS + P,
\end{equation}
where $J = \partial f/\partial X$, and $P = \partial f/\partial \theta$.

The Cramer-Rao bound states that the inverse of the FIM provides
a lower bound of parameter estimation variance-covariance matrix, which is the fundamental support for FIM-based OED.
The target of OED is to optimize a scalar measure of FIM using different criteria. The objective function of OED can be expressed as
\begin{equation}\label{timlambda}
	\lambda^* = \arg \min_{{\lambda} \in \Omega} \Phi \left( \left( \text{FIM}(\theta, \lambda) \right)^{-1} \right)
\end{equation}
where $\lambda$ is a vector representing design factors and $\Omega$ is the possible space for $\lambda$. FIM is characterized by both the model parameters $\theta$ and the design factors $\lambda$. $\Phi(\cdot)$ denotes the design
criteria such as A-, D- and E- optimal design that will get scalar features from FIM \cite{hosten1974sequential}.  In this paper, we will focus on the E-optimal method.

According to Martin Brown \cite{brown2008robust}, E-optimal design can be cast as a Semi-Definite
Programme
\begin{equation}\label{optimal:toiuuloi}
	\begin{array}{ll}
		\text{min} & -t \\
		\text{s.t.} & \sum_{i=1}^{N}\lambda_{i}S^{T}_{i}S_{i}\succeq t{\bf I}\\
		& \lambda_{i}\succeq 0, \forall i\\
		& {\bf 1}^{T}\lambda = 1,
	\end{array}
\end{equation}
where $S_i=\partial X/\partial \theta_i$  is an $i$-th column vector of $S$, ${\bf I}$ is an identity matrix.
Transforming an E-optimal problem into a convex optimization problem offers several significant advantages, primarily due to the inherent properties of convex optimization problems and the corresponding solution techniques. 
Convex optimization problems exhibit global convergence properties, ensuring that solution methods typically converge to a global optimum rather than getting trapped in local optima, as can occur in non-convex problems. Additionally, numerous powerful optimization tools and software, such as CVXPY, MOSEK, Gurobi, and optimization libraries in Python like \texttt{scipy.optimize}, are specifically designed for convex problems, facilitating easier and more convenient implementation and resolution. In this work, we use the CVXPY library \cite{Boydl2019dgp} to solve the Semi-Definite Programme (\ref{optimal:toiuuloi}).

\section{Simulation-based Methods for Optimal Sampling Design}

In this paper, we propose two simulation-based methods to find optimal sampling times. Both methods do not require an initial estimate of the parameters. 
The first method, called E-optimal-ranking (EOR), combines simulated parameters and E-optimal criterion. The second method combines simulated data and At-LSTM.

\subsection{E-optimal-ranking (EOR)}
We propose a new simulation-based method for optimal sampling design that employs E-optimal criterion, called E-optimal-ranking. 
We uniformly select parameters within the parameter space $\Theta$ to solve the optimization Problem \eqref{optimal:toiuuloi}. Its solution, $\lambda_i$, is collected as the basis for ranking the time points $t_i$. The ranking is from $1$ to $N$ where $1$ is the highest rank and $N$ is the lowest rank. If $\lambda_i > \lambda_j$, then $t_i$ ranks higher than $t_j$.
We repeat the above process with a sufficiently large number of times.  
Then, the average rank of each time point is computed.
We choose $n$ optimal sampling times by taking the $n$ highest ranked time points.
The method is described in Algorithm \ref{alg:eor_algorithm}.
This method is more flexible than the classical E-optimal criterion because it does not require an initial estimate of the parameters.

\begin{algorithm}
\caption{EOR}
\label{alg:eor_algorithm}
\begin{algorithmic}[1]
    \REQUIRE System of ODEs (1), (5); parameters space $\Theta$; set of time points $(t_i)_{i=1}^N \subset [0, T]$; $K$  is a number of times for solving Problem \eqref{optimal:toiuuloi}.
    \ENSURE Set of $n$ optimal time points is a subset of $(t_i)_{i=1}^N$.

    \FOR{$j = 1$ \TO $K$}
        \STATE Sample uniformly $\theta \in \Theta$. 
        \STATE Solve system (1), (5).
        \STATE Constructing FIM.
        \STATE Solve convex optimization Problem \eqref{optimal:toiuuloi}.
        \STATE Collect $\lambda_i$ and rank $(t_i)_{i=1}^N$ based on $\lambda_i$, $i = 1,\ldots, N$. 
    \ENDFOR

    \STATE Calculate the average ranking of $(t_i)_{i=1}^N$.
    \STATE Determine the optimal time points with the highest average ranking.

    \RETURN Set of $n$ optimal time points
\end{algorithmic}
\end{algorithm}

\subsection{Attention-based Long short-term memory (At-LSTM)}

Wang et al. \cite{wang2016attention} proposed a unique idea of incorporating the attention mechanism into Long Short-Term Memory (LSTM) models to enhance prediction accuracy. This attention mechanism in LSTM helps us identify important features by assigning attention weights to the data features. In particular, we first uniformly select parameters from the parameter space $\Theta$.
For each set of simulated parameters, we simulate data at the possible observational times according to Equation (\ref{eqs:hedonghoctongquat}).
Then, we build an At-LSTM model to map each data to the corresponding parameters.
Attention weights are then extracted at each time point. 
Finally, the time points with the highest attention weights are selected as the optimal sampling times (see Algorithm \ref{alg:lstm_attention} for details).

\begin{algorithm}
\caption{At-LSTM Model for Optimal Sampling Time}
\label{alg:lstm_attention}
\begin{algorithmic}[1]
    \REQUIRE System of ODEs \eqref{eqs:hedonghoctongquat}; the space $\Theta$ contain $\theta$ ; set of time points $(t_i)_{i=1}^N \subset [0, T]$; $K$ is a number of observations of dataset.
    \ENSURE Set of $n$ optimal time points
    \FOR{$j = 1$ \TO $K$}
        \STATE Sample uniformly $\theta \in \Theta$. 
        \STATE Simulate data at $(t_i)_{i=1}^N$ according to $\theta$.
    \ENDFOR
    \STATE Train an At-LSTM model to map data to the corresponding $\theta$.
    \STATE Extract attention weights. 
    \RETURN Set of $n$ optimal time points is the set that has the highest weights.
\end{algorithmic}
\end{algorithm}


\section{Case study}

\begin{figure}[htbp]
	\centering
	\includegraphics[width=\columnwidth]{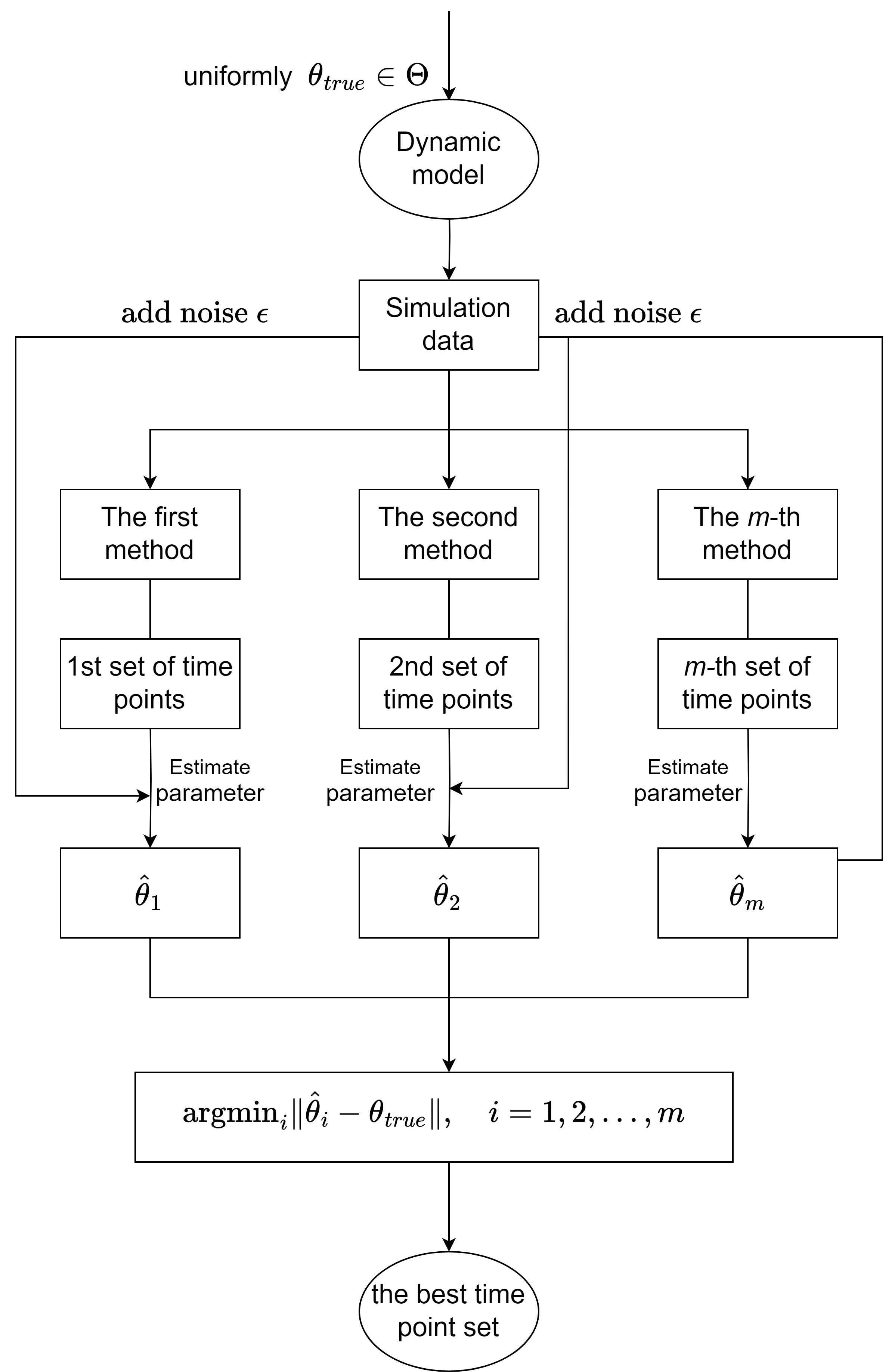}
	\caption{The comparison procedure between several optimal sampling schemes.}
	\label{fig:quy trinh}
\end{figure}

In this section, we use simulations to compare the performance of our proposed methods with the E-optimal criterion and a naive method where sampling points are chosen randomly.
Specifically, we simulate data from two popular models and apply several methods to identify the optimal sampling points.
 The set of time points that results in the lowest estimated error of the parameter will be considered the best. This process is illustrated in Figure \ref{fig:quy trinh}.

\subsection{Lotka - Volterra model}

The Lotka–Volterra dynamic model, known as the prey-predator model, describes the interaction between two competing species.  The model was independently developed by  the American biophysicist Alfred J. Lotka in 1925 \cite{lotka1925elements} and the Italian mathematician Vito Volterra in 1927 \cite{volterra1927variazioni}. This is an important tool for studying biological populations in a natural environment.
The Lotka-Volterra Model has many applications in various fields including chemistry \cite{Chemical1989mathematical}, ecology \cite{Ecology1974population}, biology \cite{hernandez1997lotka}, business \cite{Economic2021competition}, and epidemiology \cite{mohammed2021analytical}.

To generate simulated data, we solve the following ODE system
\begin{equation}\label{LV_system}
	\left\{
	\begin{aligned}
		\frac{dr}{dt} &= \alpha r - \beta rw, \\
		\frac{dw}{dt} &= -\gamma w + \delta rw.
	\end{aligned}
	\right.
\end{equation}
We follow the the paper \cite{nguyen2019active} and use the parameter space $(\alpha, \beta, \gamma, \delta) \in \Theta  = [0.5, 1.5] \times [0.01, 0.1] \times [0.5, 1.5] \times [0.01, 0.1]$ and 50 as the initial value of both predator and prey populations. 
Possible observation time points are $101$ points that are placed equally in the time interval $[0, 10]$.
For the comparison, we simulate $1000$ observations by first simulating $1,000$ values of $(\alpha, \beta, \gamma, \delta)$ uniformly in $\Theta$, and then for each simulated parameter set, we simulate the observations of $w$ at the possible time points. For each observation, we apply the four optimal design sampling methods for selecting 5 best sampling time points to estimate the coefficients of the system of Equations \eqref{LV_system} as follows:

\subsubsection{Random} We randomly select $5$ time points from the total of $101$ time points. This method provides different time points for different datasets.

\subsubsection{E-optimal} For each simulated dataset, we use the true parameters of that dataset as the initial estimate and solve problem \eqref{optimal:toiuuloi}. 
Then, the top 5 time points with the highest values $\lambda_i$ are selected as the optimal sampling times.
Again, this method gives different sets of optimal times for different datasets.

\subsubsection{EOR} We perform Algorithm \ref{alg:eor_algorithm} with $N = 101, n = 5, T =10, K=1,000$. Note that this algorithm provides one set of best sampling times for all datasets. The set of optimal time points is $\{2.1, 2.2, 2.3, 2.4, 2.5\}.$

\subsubsection{At-LSTM} We use Algorithm \ref{alg:lstm_attention} with $N = 101, n = 5, T =10, K=100,000$. As a result, the only set of optimal time points for all simulated datasets is $\{1.0, 1.1, 1.2, 1.3, 1.4\}.$ 

\begin{figure}[htbp]
	\centering
	\includegraphics[width=\columnwidth]{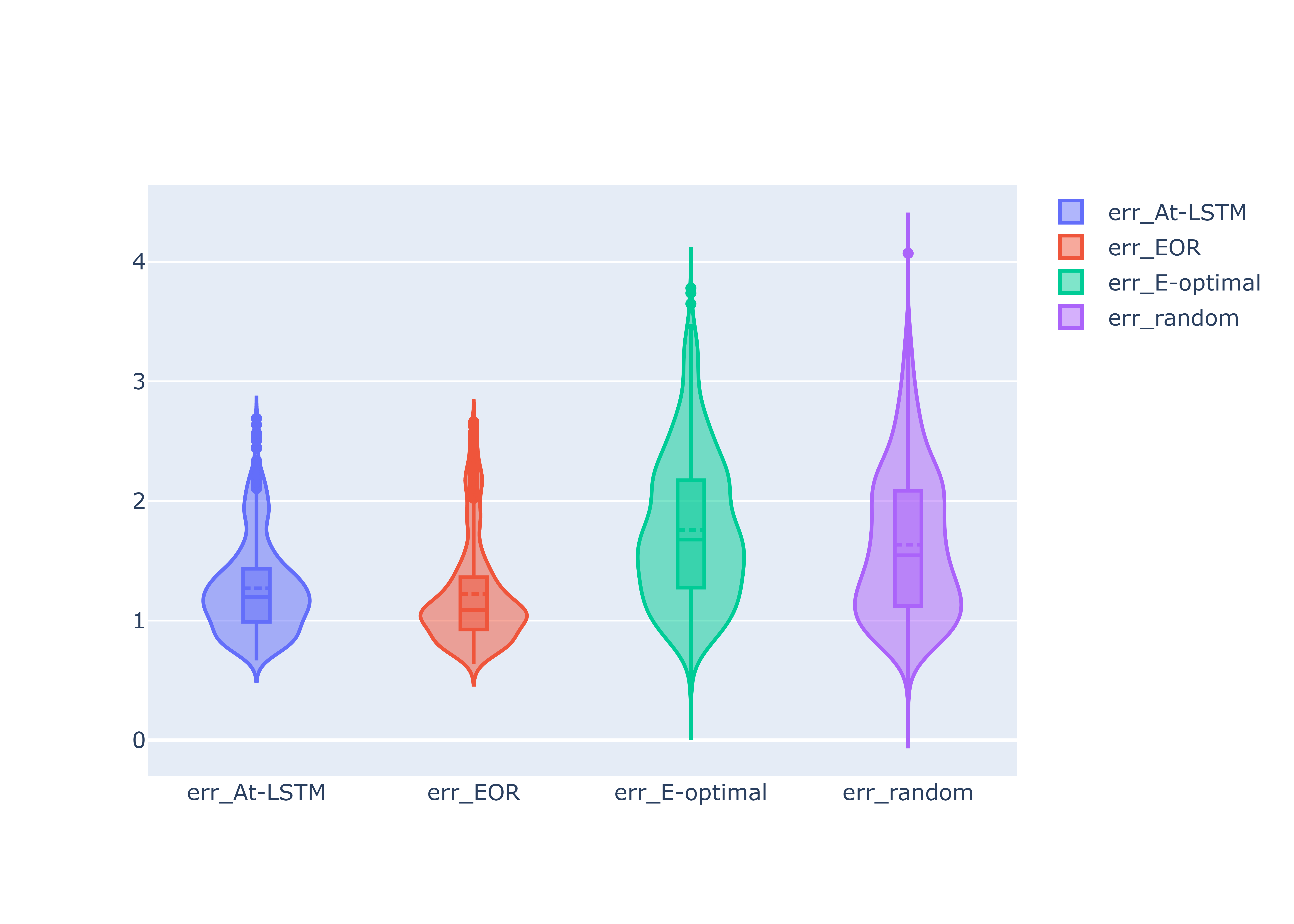}
	\caption{Violine plots of the average error of four optimal sampling design methods under the Lotka-Volterra model.}
	\label{fig:LV}
\end{figure}

To determine which sampling method is best for estimating the parameters of the dynamic system, we use the average error of the least squares estimate \eqref{LS} with respect to observations at the optimal time points selected by each method.  
The results of this process are presented in Table \ref{tab:combined-error-summary} and Figure \ref{fig:LV}.
We can see that our proposed methods (EOR and At-LSTM) perform significantly better than random selection and E-optimal criterion.

\begin{table*}[htbp]
	\centering
	\caption{Summary of the average error of four optimal sampling design methods under the Lotka-Volterra and three-compartment pharmacokinetic models.}
	\label{tab:combined-error-summary}
	\begin{tabular}{|lllll|llll|}
		\hline
		{} & \multicolumn{4}{c|}{Lotka-Volterra Model} & \multicolumn{4}{c|}{Three-Compartment Pharmacokinetic Model} \\
		\hline
		{} & At-LSTM & EOR & E-optimal & Random & At-LSTM & EOR & E-optimal & Random \\
		\hline
		Mean  & 1.27 & 1.22 & 1.76 & 1.63 & 0.77 & 0.55 & 0.55 & 1.08 \\
		Std   & 0.39 & 0.44 & 0.61 & 0.61 & 0.50 & 0.26 & 0.27 & 0.61 \\
		Min   & 0.67 & 0.63 & 0.34 & 0.27 & 0.06 & 0.05 & 0.03 & 0.02 \\
		25\%   & 0.99 & 0.93 & 1.28 & 1.12 & 0.45 & 0.36 & 0.34 & 0.57 \\
		50\%   & 1.20 & 1.09 & 1.68 & 1.55 & 0.67 & 0.55 & 0.53 & 1.03 \\
		75\%   & 1.43 & 1.36 & 2.17 & 2.09 & 0.95 & 0.73 & 0.71 & 1.50 \\
		Max   & 2.69 & 2.66 & 3.78 & 4.07 & 3.83 & 1.99 & 2.27 & 3.31 \\
		\hline
	\end{tabular}%
\end{table*}

We perform Tukey's HSD test to confirm the statistically significant differences between the performance of these sampling methods. To test the differences between statistics, researchers use the family-wise error rate (FWER) as the probability of making a Type I error among a specified group of tests. The results of Tukey's HSD test with FWER = 0.05 are presented in Table \ref{tab:combined-Tukey-test}.
They conclude that EOR and At-LSTM are significantly better than Random selection and E-optimal methods.
On the other hand, the performance of EOR and At-LSTM is similar.

\begin{table*}[htbp]
    \centering
    \caption{Comparing the performance of four optimal sampling design methods under the Lotka-Volterra and three-compartment pharmacokinetic models using Tukey's HSD test with FWER=0.05.}
    \label{tab:combined-Tukey-test}
    \begin{tabular}{|l l c l c c c|l c l c c |}
        \hline
        \multicolumn{7}{|c|}{Lotka-Volterra Model} & \multicolumn{5}{|c|}{Three-Compartment Pharmacokinetic Model} \\
        \hline
        Group 1 & Group 2 & Meandiff & p-adj & Lower & Upper & Reject & Meandiff & p-adj & Lower & Upper & Reject \\
        \hline
        E-optimal & EOR & -0.53 & 0.0 & -0.60 & -0.47 & True & 0.01 & 0.99 & -0.04 & 0.06 & False \\
        E-optimal & At-LSTM & -0.49 & 0.0 & -0.55 & -0.43 & True & 0.23 & 0.0 & 0.18 & 0.28 & True \\
        E-optimal & Random & -0.12 & 0.0 & -0.19 & -0.06 & True & 0.53 & 0.0 & 0.48 & 0.58 & True \\
        EOR & At-LSTM & 0.05 & 0.26 & -0.02 & 0.11 & False & 0.22 & 0.0 & 0.17 & 0.27 & True \\
        EOR & Random & 0.41 & 0.0 & 0.35 & 0.47 & True & 0.53 & 0.0 & 0.48 & 0.57 & True \\
        At-LSTM & Random & 0.36 & 0.0 & 0.30 & 0.43 & True & 0.30 & 0.0 & 0.25 & 0.35 & True \\
        \hline
    \end{tabular}
\end{table*}


\subsection{Three-compartment pharmacokinetic model}

The three-compartment pharmacokinetic model is a system of first-order linear differential equations widely used to describe the absorption, metabolism, and clearance of drugs in the human body. 
It divides the body into three compartments: the central compartment, which typically symbolizes blood and well-perfused organs; and two peripheral compartments, representing place tissues.
The drug may be stored or metabolized more slowly. The three-compartment model is widely applied in clinical pharmacology and drug development for predicting drug behavior \cite{derendorf2000pharmacokinetic}, optimizing medication regimens \cite{gerlowski1983physiologically}, and ensuring safe and effective treatment \cite{garrett2018pharmacokinetic}. It also helps to understand how dangerous compounds are distributed in the body. From there, the treatment strategy may be tailored to each individual.

We simulate data according to the following three-compartment pharmacokinetic model:
\begin{equation}\label{3_ngan}
	\left\{
	\begin{aligned}
		\frac{dx_1(t)}{dt} &= -(k_{10} + k_{12} + k_{13})x_1(t) + k_{21}x_{21}(t) + k_{31}x_3(t), \\
		\frac{dx_2(t)}{dt} &= k_{12}x_1(t) - k_{21}x_2(t), \\
		\frac{dx_3(t)}{dt} &= k_{13}x_1(t) - k_{31}x_3(t),
	\end{aligned}
	\right.
\end{equation}
We follow the paper \cite{kartono2021study} and use the parameter space  $(k_{10},k_{12},k_{13}, k_{21}, k_{31}) \in \Theta = [0.09, 2.40] \times  [0.5, 1.0] \times [0.5, 2.0] \times [0.2, 0.6]\times [0.5, 0.7]$. The initial values of the drug amounts in compartments $C_1$, $C_2$, and $C_3$ are 100, 0, and 0, respectively. We consider $101$ possible observation time points that are divided equally in the time interval [0, 25]. 
We simulate 1000 datasets by first simulating 1000 values of $( k_{10},k_{12},k_{13}, k_{21}, k_{31})$ uniformly in $\Theta$, and then simulate the observations of $C_1$ (that is $x_1$) at the possible observation time points for each set of simulated parameters. 
We apply four optimal design sampling methods for selecting 5 best
sampling time points as follows:
  
 \subsubsection{Random} We randomly select $5$ time points out of a total of 101 time points. 
 
 \subsubsection{E-optimal} For each parameter $\theta$, we solve Problem \eqref{optimal:toiuuloi} in the case of the three-compartment model, System \eqref{3_ngan}.  Then, the top 5 time points with the highest values $\lambda_i$ are selected to estimate the coefficients of the system of equations.

\subsubsection{EOR} We perform Algorithm \ref{alg:eor_algorithm} with $N = 101, n = 5, T = 25, K = 1,000$. As a result, we obtain  \{0.25, 0.5, 3.75, 4.0, 4.25\} as the optimal sampling times for all datasets.

\subsubsection{At-LSTM} We use Algorithm \ref{alg:lstm_attention} with $N = 101, n = 5, T = 25, K = 100,000$. 
The set of optimal sampling times for all datasets is $\{1.0, 1.25, 3.5, 3.75, 3.25\}$.

\begin{figure}[htbp]
	\centering
	\includegraphics[width=\columnwidth]{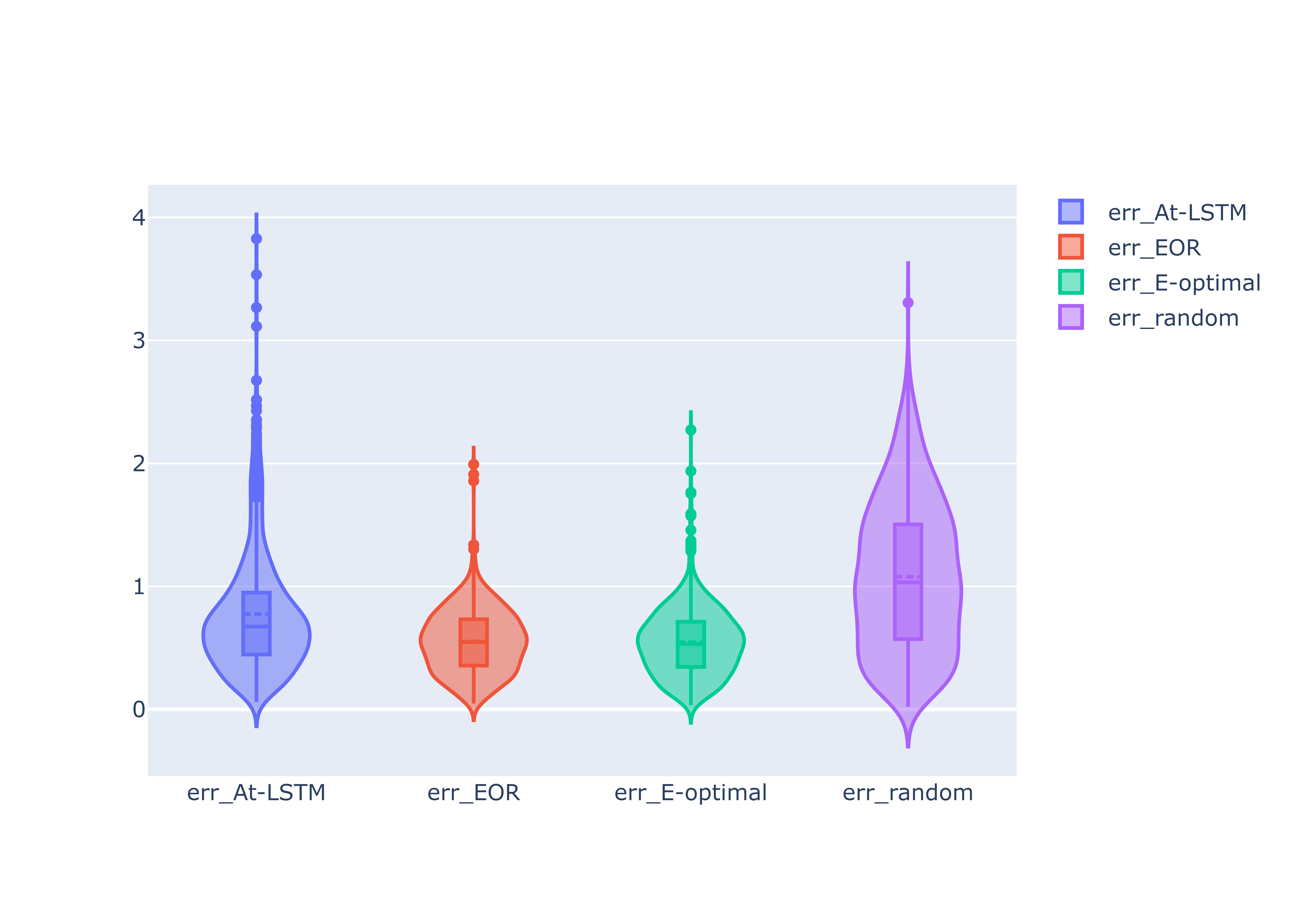}
	\caption{Violine plots of the average error of four optimal sampling design methods under the three-compartment model.}
	\label{fig: so sanh 3 ngan}
\end{figure}

Similar to the previous case study, we compare the average errors of all methods. 
The results of the average errors are presented in Table \ref{tab:combined-error-summary} and Figure \ref{fig: so sanh 3 ngan}.
In this case, we see that the performance of the EOR and E-optimal are similar and better than random selection and At-LSTM.
Tukey's HSD test with FWER = 0.05 confirms this conclusion (see Table \ref{tab:combined-Tukey-test}).

\section{Discussion and Conclution}

Choosing optimal sampling points for estimating parameters of dynamical systems is a critical step in studying biological systems. 
In this paper, we present two new simulation-based strategies for selecting optimal sampling times (EOR and At-LSTM). 
Unlike classical optimal sampling methods, our approaches avoid the need for an initial estimate of the parameters.
We demonstrate that our proposed methods are better than the classical E-optimal criterion.
However, we note that the performance of At-LSTM depends heavily on the performance of the LSTM neural network, which is affected by tuning.
Moreover, if we replace LSTM with other machine learning models in Algorithm \ref{alg:lstm_attention}, we will obtain new methods.
One future direction is to study the impact of using different machine learning models in Algorithm \ref{alg:lstm_attention}.
Another direction is to explore the possibility of combining several optimal sampling schemes using ensemble methods.

\section*{Acknowledgment}
LSTH was supported by the Canada Research Chairs program, the NSERC Discovery Grant RGPIN-2018-05447, and the NSERC Discovery Launch Supplement DGECR-2018-00181.
We would like to extend our sincere gratitude to the Scientific Research Fund of Saigon Technology University for their generous support and funding.

\bibliographystyle{IEEEtran}
\bibliography{TrichDan}

\end{document}